# Analysis of the Arabic Broken Plurals and Diminutive

George Anton Kiraz*
University of Cambridge (St John's College)
Computer Laboratory
Pembroke Street, Cambridge CB2 1TP
George.Kiraz@cl.cam.ac.uk
URL: http://www.cl.cam.ac.uk/users/gk105/

**Abstract**

This paper demonstrates how the challenging problem of the Arabic broken plurals and diminutive can be handled under a multi-tape two-level model, an extension to two-level morphology.

## 1  Introduction

The phenomenon of the broken plural and diminutive in Arabic poses a challenge to main-stream two-level morphology, not only because of its nonconcatenative nature, but also because its analysis relies heavily on prosodic structure. The purpose of this paper is to present an implemented morphological model which is capable of analysing the broken plural.

The following convention has been adopted. Morphemes are represented in braces, { }, and surface forms in solidi, / /. In examples of grammars, variables begin with a capital letter. Cs denote consonants and Vs denote vowels. In two-level rules, square brackets mark optional segments.

The structure of the paper is as follows: section 2 presents the problem of the broken plural; section 3 introduces a computational framework for solving the problem; section 4 demonstrates how the broken plural may be derived via the 'implicit derivation' of the singular under a two-level model; finally, section 5 gives concluding remarks.

## 2  Problem Description

The derivation of the broken plurals and diminutive in Arabic is a complicated task. Consider the data in (1) – from McCarthy and Prince (1990a).

(1)  BROKEN PLURAL AND DIMINUTIVE DATA

| Singular | Plural | Diminutive | Gloss |
|---|---|---|---|
| ĵundub | ĵanaadib | ĵunaydib | 'locust' |
| sulṭaan | salaaṭiin | sulayṭaan | 'sultan' |

The analysis of the plural makes use of prosodic structure, where a word (W) consists of at least one foot (F), and feet consist of syllables ($\sigma$) as in (2).

---

*Supported by a Benefactor Studentship from St John's College. This research was done under the supervision of Dr Stephen G. Pulman.

(2) PROSODIC HIERARCHY

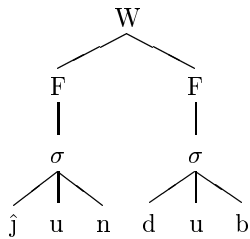

McCarthy and Prince (1990a) show that broken plurals have the plural template in (3).[1]

(3) PLURAL TEMPLATE

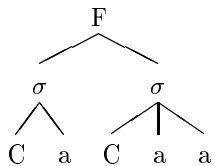

The derivation of the plural from the singular takes the following path. The singular (e.g. /ĵundub/) is factored into two partitions: a **kernel** which consist of the first foot (i.e. /ĵun/) and a **residue** which is the remainder (i.e. /dub/). The kernel /ĵun/ is mapped onto the plural template: the consonants link to the Cs, and the default plural vowel [a] remains. This results in the representation in (4).

(4) DERIVING /ĵanaadib/

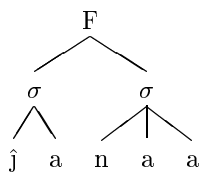

Then, the residue /dub/ is added. This is straightforward, except that the default vowel for the residue is [i] (i.e. /dub/ becomes /dib/) resulting in /ĵanaadib/. The derivation of /salaaṭiin/ from /sulṭaan/ follows the same path. The kernel /sul/ maps to the plural template producing /salaa/, and the residue /ṭaan/ is added with the vowel [i]; hence, /salaaṭiin/. The diminutive can be derived in a similar fashion.

Traditional grammars provide a much simpler analysis. Wright (1988) lists 31 broken plural patterns. Each root must be lexically marked with its own pattern. For example, the root {ĵndb} is marked with the pattern CaCaaCiC, and {slṭn} is marked with CaCaaCiiC. Instantiating the Cs with the root consonants produces /ĵanaadib/ and /salaaṭiin/, respectively. The analysis of McCarthy and Prince, however, provides a wider generalisation: the operation of **pluralisation**, which can be achieved by morphological rules in a computational framework, rather than by explicit, manual marking of lexical entries.

# 3 Computational Framework

This section presents a computational framework under which the broken plural can be analysed. Section 3.1 gives some background to Arabic computational morphology, and section 3.2 defines our model.

## 3.1 Background

Two-level morphology (Koskenniemi 1983) defines two levels of strings in recognition and synthesis: lexical strings represent morphemes, and surface strings represent surface forms. Two-level rules map the two strings; the rules are compiled into finite state transducers, where lexical strings sit on one tape of the transducers and surface strings on the other.

---

[1] The default plural vowel melody of the broken plural is {ai} except in CVCC and CVCVC singular stems.

Traditional two-level models are biased towards concatenative morphology since they assume that the lexical string is the concatenation of lexical morphemes. This requirement makes it extremely difficult, if not impossible, to analyse the data presented in section 2.

A number of proposals for handling Arabic morphology exist[2]; none provide a framework under which the phenomenon of the broken plural can be analysed in a linguistically motivated method. An implemented system which can handle the broken plural is the ALPNET system reported by (Beesley et al. 1989; Beesley 1990; Beesley 1991). In this system, root entries in the lexicon are associated with a set of nominal patterns, some of which indicate the broken plural (Beesley, *personal communication*). For example, the root {ĵ_n_d_b} (where _ represents an infix) is associated with the patterns {_u__u_} 'singular' and {_a__aa_i_} 'plural' (where _ represents a root consonant). The intersection of the root morpheme and pattern morphemes produces /ĵundub/ and /ĵanaadib/, respectively.

There is no doubt that the ALPNET system can produce broken plurals. However, its handling of the broken plural follows traditional grammars (the ALPNET project started before the new findings of – see section 2 – were published). It is more desirable to have a morphological model which can capture the generalisations in these findings. Under this framework, the operation of 'pluralisation' would be handled by two-level rules, rather than marking each root entry in the lexicon with its broken plural pattern(s). Hence, lexicon maintenance would be easier and more efficient, specially that pluralisation is productive in Arabic, e.g. /ṣandal/ ← /ṣanaadil/ 'sandal'. The next subsection introduces such a model.

## 3.2 A Computational Model

There are two main difficulties in handling broken plurals. The first is choosing a formalism which allows the user to write two-level rules for deriving broken plurals along the lines described in section 2. The formalism must allow the mapping between lexical and surface strings of *unequal* lengths. For example, in the derivation of /ĵanaadib/ from /ĵundub/, one needs to map the kernel /ĵun/ to /ĵanaa/. A formalism which fits this criterion appears in (5) (Ruessink 1989; Pulman and Hepple 1993).[3]

(5) TWO-LEVEL FORMALISM
   a. LSC - SURF - RSC ⇒ LLC - LEX - RLC
   b. LSC - SURF - RSC ⇔ LLC - LEX - RLC
   where
   | LSC | = | left surface context | LLC | = | left lexical context |
   |---|---|---|---|---|---|
   | SURF | = | surface form | LEX | = | lexical form |
   | RSC | = | right surface context | RLC | = | right lexical context |

The special symbol * indicates an empty context, which is always satisfied. The operator ⇒ states that LEX *may* surface as SURF in the given context, while the operator ⇔ adds the condition that when LEX appears in the given context, then the surface description *must* satisfy SURF. The latter caters for obligatory rules. A lexical string maps to a surface string iff they can be partitioned into pairs of lexical-surface subsequences, where each pair is licenced by a rule. Rules are associated with a feature structure which must unify with the lexical feature structure of the morpheme affected by the rule.

In order to illustrate how the formalism can be used for deriving broken plurals, let us ignore for the moment the templatic nature of Arabic morphology. Assume that the lexicon maintains the singular stem /ĵundub/. Two rules are required for the derivation of the plural /ĵanaadib/ (6).

---

[2] Previous works include: Kay (1987), Kornai (1991), Wiebe (1992), Narayanan and Hashem (1993), Pulman and Hepple (1993), and Bird and Ellison (1994).

[3] Our implementation interprets rules directly; hence, we allow unequal representation of strings. If the rules were to be compiled into automata, a genuine symbol, e.g. 0, must be introduced by the rule compiler. For the compilation of our formalism into automata, see Grimley-Evans, Kiraz & Pulman, *forthcoming*.

(6) TWO-LEVEL RULES

R1: $\begin{array}{ccccccc} * & - & C_1V_1C_2 & - & C_3V_2[V_2]C_4 & \Leftrightarrow \\ * & - & C_1aC_2aa & - & * & \end{array}$

R2: $\begin{array}{ccccccc} C_1V_1C_2 & - & C_3V_2C_4 & - & * & \Leftrightarrow \\ * & - & C_3iC_4 & - & * & \end{array}$

R3: $\begin{array}{ccccccc} C_1V_1C_2 & - & C_3V_2V_2C_4 & - & * & \Leftrightarrow \\ * & - & C_3iiC_4 & - & * & \end{array}$

R1 maps the kernel /ĵun/ to /ĵanaa/ and R2 maps the residue /dub/ to /dib/. R3 is similar to R2 except that it sanctions residues with a long vowel, e.g. /ṭaan/ in /sulṭaan/. The lexical contexts ensure that the proper rule is applied to the kernel, B:Φ, or the residue, B/Φ. (7) illustrates the two-level derivations of /ĵanaadib/ and /salaaṭiin/.

(7) TWO-LEVEL ANALYSIS OF /ĵanaadib/ AND /salaaṭiin/

| ĵun | dub | | sul | ṭaan | *Lexical Tape* |
|-----|-----|---|-----|------|----------------|
| R1  | R2  | | R1  | R3   |                |
| ĵanaa | dib | | salaa | ṭiin | *Surface Tape* |

The second difficulty in analysing the broken plural is a result of the nature of Arabic morphology, where the majority of Arabic stems are templatic, i.e. they are derived from a root and a vowel melody according to a specific template (or pattern). For example, the singular /ĵundub/ is derived from the root morpheme {ĵndb} and the vowel melody morpheme {u}; both are arranged according to the template morpheme {CVCCVC}. This derivation is illustrated in (8).[4]

(8) DERIVATION OF /ĵundub/

$$/\hat{\jmath}undub/ = \begin{array}{c} \quad\quad u \\ \quad\quad | \\ C\,V\,C\,C\,V\,C \\ |\quad\;\;|\;\;|\quad\;\;| \\ \hat{\jmath}\quad n\,d\quad b \end{array}$$

Kiraz (1994) proposed some extensions to the above formalism in order to handle Arabic templatic morphology. One of the extensions introduced is that all expressions in the lexical side of the rules (i.e. LLC, LEX and RLC) are $n$-tuple of regular expressions of the form $(x_1, x_2, \ldots, x_n)$. The $i$th expression refers to symbols on the $i$th tape. When $n = 1$, the parentheses can be ignored; hence, $(x)$ and $x$ are equivalent; a nil slot is indicated by $\varepsilon$. (The original idea of using multiple tapes is due to Kay (1987).)

Assuming that the lexicon maintains the roots {ĵndb} and {slṭn} (each root is associated with the feature [number=N]), the vocalisms {uu} and {uaa},[5] and the patterns {cvccvc} and {cvccvvc}, the derivation of singular stems can be achieved by the rules in (9).

(9) TWO-LEVEL RULES

R4: $\begin{array}{ccccccc} * & - & (c, X, \varepsilon) & - & * & \Rightarrow & \text{FEATURES: [number=sing]} \\ * & - & X & - & * & \end{array}$

R5: $\begin{array}{ccccccc} * & - & (v, \varepsilon, X) & - & * & \Rightarrow & \text{FEATURES: [number=sing]} \\ * & - & X & - & * & \end{array}$

R4 sanctions consonants and R5 sanctions vowels. The two-level derivation of /ĵundub/ and /sulṭaan/ appears in (10).

---

[4]The template morpheme is presented here in CV terms in order to simplify the presentation and in order to concentrate on the broken plural issue. Templates can be specified by prosodic terms such as syllable and mora (McCarthy and Prince 1990b; McCarthy 1993).

[5]For simplicity, the vocalism {u} and {ua} are entered as above to avoid writing spreading rules; for handling spreading, see Kiraz (1994).

(10) TWO-LEVEL ANALYSIS OF /ĵundub/ AND /sulṭaan/

| u | | u | | | Vocalism Tape (VT) |
|---|---|---|---|---|---|
| ĵ | n | d | | b | Root Tape (RT) |
| c | v | c | c | v | c | Pattern Tape (PT) |
| R4 | R5 | R4 | R4 | R5 | R4 | |
| ĵ | u | n | d | u | b | Surface Tape (ST) |

| u | | | a | a | | Vocalism Tape (VT) |
|---|---|---|---|---|---|---|
| s | | l | ṭ | | | n | Root Tape (RT) |
| c | v | c | c | v | v | c | Pattern Tape (PT) |
| R4 | R5 | R4 | R4 | R5 | R5 | R4 | |
| s | u | l | ṭ | a | a | n | Surface Tape (ST) |

So far, we have seen how it is possible to derive plurals from singular stems, and how to derive singular stems from morphemes (i.e. roots, vocalisms and patterns). The next section looks into how this model is capable of deriving plural forms from morphemes, rather than from singular stems.

## 4 Analysis of the Broken Plural

Because of its dependency on the singular (which does not constitute a lexical entry), the broken plural seems to require a *three*-level derivation. In other words, the derivation of the broken plural takes the following form: (root, vocalism, pattern) → singular → plural. There are two methods for deriving the plural (the choice depends on the rest of the grammar). The first requires a *three*-level model as illustrated in (11a); it is possible to collapse the three representations into two via composition. The second derives the plural directly from the root, pattern and vocalism morphemes via the **implicit derivation** of the singular using just our two-level model. The concept is illustrated in (11b).

(11) TWO-LEVEL VS THREE-LEVEL DERIVATION
    a. Three level System      b. Two-level System

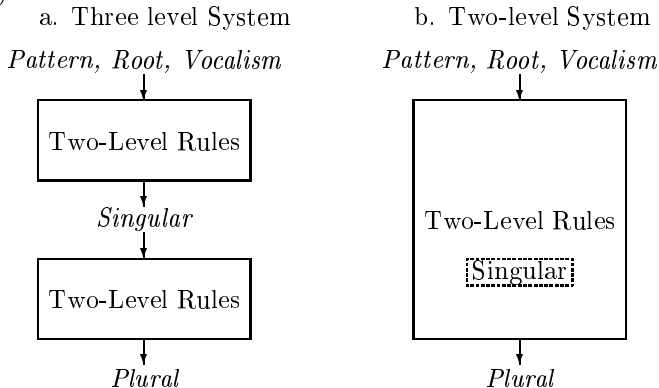

Deriving the broken plural via the 'implicit derivation' of the singular is best explained by an example. To derive /ĵanaadib/ (singular /ĵundub/), a two-level rule scans the kernel, i.e. /ĵun/, from lexical morphemes; however, instead of mapping the kernel to /ĵun/ on the surface, the rule maps it to the plural template CaCaa, i.e. /ĵanaa/. A second rule maps the rest of the morpheme characters, i.e. the residue /dub/, to the surface as they are, but overwriting the singular vowel melody with the plural one. The derivation, then, remains within two-level theory. We say that the plural is derived via the 'implicit derivation' of the singular because the two-level rules find a singular form, but map it to the corresponding plural form on the surface. To illustrate this process, consider the rules in (12).

(12) TWO-LEVEL RULES

R6: $\begin{matrix} * & - & (cvc, C_1C_2, V_1) & - & (cv[v]c, C_3C_4, V_2[V_2]) & \Leftrightarrow \\ * & - & C_1aC_2aa & - & * & \end{matrix}$

R7: $\begin{matrix} (cvc, C_1C_2, V_1) & - & (cvc, C_3C_4, V_2) & - & * & \Leftrightarrow \\ * & - & C_3iC_4 & - & * & \end{matrix}$

R8: $\begin{matrix} (cvc, C_1C_2, V_1) & - & (cvvc, C_3C_4, V_2V_2) & - & * & \Leftrightarrow \\ * & - & C_3iiC_4 & - & * & \end{matrix}$

R6-R8 perform the same thing as R1-R3, except that (1) they operate on three lexical tapes instead of one, and (2) they are all marked with the feature [number=pl]. The two-level derivation of /ĵanaadib/ and /salaaṭiin/ under this scheme is illustrated in (13).

(13)  TWO-LEVEL ANALYSIS OF /ĵanaadib/ AND /salaaṭiin/

| u | u |   | u | aa | $VT$ |
|---|---|---|---|----|------|
| ĵn | db |  | sl | ṭn | $RT$ |
| cvc | cvc |  | cvc | cvvc | $PT$ |
| R6 | R7 |  | R6 | R8 |      |
| ĵanaa | dib |  | salaa | ṭiin | $ST$ |

A few points should be noted: Firstly, R6-R8 are implicitly finding a singular derivation from the lexicon, but mapping such derivation to the corresponding plural form on the surface; note that the vowels which appear on VT are those of the singular /ĵundub/ and /sulṭaan/. Secondly, the rules are all obligatory. Thirdly, the rule feature structure `[number=pl]` must unify with the lexical feature structure `[number=N]`. Finally, the lexical structures in (13) and (10) are equivalent.

Since Vs on the VT tape can unify with any vowel in the lexicon, the two-level module produces many analyses depending on the number of vocalisms in the lexicon. This is solved by associating lexical entries with feature structures for morphotactic parsing. Only the analyses with the proper singular vocalic melody are parsed successfully using a unification-based morphotactic grammar (Bear 1986; Ritchie et al. 1992).

## 5   Conclusion

I have presented in this paper a computational framework which can handle Arabic broken plurals in a linguistically-motivated method. The implementation has been tested on all classes of stems, even those which require the insertion of a default consonant [w], e.g. /ĵawaamiis/ 'buffaloes', and cases where this consonant is realised as [ʔ], e.g. /ĵazaaʔir/ 'islands' (Kiraz 1996). Diminutive and the phenomenon of the maṣâdir can be handled in a similar manner.